\documentclass[conference]{IEEEtran}
\usepackage{xcolor}
\IEEEoverridecommandlockouts

\makeatletter
\newcommand{\linebreakand}{%
  \end{@IEEEauthorhalign}
  \hfill\mbox{}\par
  \mbox{}\hfill\begin{@IEEEauthorhalign}
}
\makeatother
\usepackage{enumitem}

\usepackage{cite}
\usepackage{amsmath,amssymb,amsfonts}
\usepackage{algorithmic}
\usepackage{graphicx}
\usepackage{textcomp}
\usepackage{xcolor}
\def\BibTeX{{\rm B\kern-.05em{\sc i\kern-.025em b}\kern-.08em
    T\kern-.1667em\lower.7ex\hbox{E}\kern-.125emX}}
\begin{document}

\title{SWMLP: Shared Weight Multilayer Perceptron for Car Trajectory Speed Prediction using Road Topographical Features\\

\thanks{

}
}

\author{
    \IEEEauthorblockN{Sarah Almeida Carneiro}
    \IEEEauthorblockA{\textit{Univ. Gustave Eiffel, LIGM}\\
                      F-77454 Marne-la-Vallée, France \\
                     \textit{IFP Energies nouvelles}\\
                     Rueil-Malmaison, France \\
                     0000-0001-7653-8614}
    \and%
    \IEEEauthorblockN{Giovanni Chierchia}
    \IEEEauthorblockA{\textit{Univ. Gustave Eiffel, CNRS, LIGM}\\
                      F-77454 Marne-la-Vallée, France \\
                      Paris, France \\
                      0000-0001-5899-689X}
    \and%
    \IEEEauthorblockN{Jean Charléty}
    \IEEEauthorblockA{\textit{Control, Signal and System Dept.} \\
                      \textit{IFP Energies nouvelles}\\
                       Rueil-Malmaison, France \\
                       }
     \linebreakand %
    \IEEEauthorblockN{Aurélie Chataignon}
    \IEEEauthorblockA{\textit{Control, Signal and System Dept.} \\
                      \textit{IFP Energies nouvelles}\\
                       Rueil-Malmaison, France \\
                       0000-0003-0112-3689}
    \and%
    \IEEEauthorblockN{Laurent Najman}
    \IEEEauthorblockA{\textit{Univ. Gustave Eiffel, CNRS, LIGM}\\
                      F-77454 Marne-la-Vallée, France \\
                      Paris, France \\
                      0000-0002-6190-0235}
}
\IEEEoverridecommandlockouts

\maketitle

\IEEEpubidadjcol

\begin{abstract}
Although traffic is one of the massively collected data, it is often only available for specific regions. One concern is that, although there are studies that give good results for these data, the data from these regions may not be sufficiently representative to describe all the traffic patterns in the rest of the world. In quest of addressing this concern, we propose a speed prediction method that is independent of large historical speed data. To predict a vehicle's speed, we use the trajectory road topographical features to fit a Shared Weight Multilayer Perceptron learning model. Our results show significant improvement, both qualitative and quantitative, over standard regression analysis. Moreover, the proposed framework sheds new light on the way to design new approaches for traffic analysis.
\end{abstract}


\section{Introduction}
\label{sec:intro}

The knowledge of a vehicle speed can substantially impact research fields influenced by traffic. The identification of daily traffic patterns, energy and pollution emission management, as well as Intelligent Transportation Systems (ITS), are some of the expertise that speed information can actively affect. 

For many years, traffic speed predictions have relied on methods such as Historical Average (HA)~\cite{smith1997traffic, kaysi1993integrated}, and Auto Regressive Integrated Moving Average (ARIMA)~\cite{smith1997traffic, billings2006application, guin2006travel}. However, traffic has a stochastic nature that involves the interconnectivity of many elements (e.g., weather, day, social events, number of cars) that could not be well represented by these models. Therefore, by using data as inputs to non-linear data-driven models, it is possible to include non-deterministic elements to the prediction and have considerable improvements. These data-driven methods exploit learning models such as: Artificial Neural Networks (ANN)~\cite{chen2012small, csikos2015traffic}; Recurrent Neural Networks (RNN)~\cite{bartlett2019prediction,vinayakumar2017applying, sadeghi2020short}; Convolutional Neural Networks (CNNs)~\cite{zhang2016dnn,zhang2017deep,song2017traffic,khajeh2019traffic,yu20193d}; and Graph Neural Networks~\cite{wang2018efficient, zhang2019graph,spinelli2020missing, ye2021spatial, zhu2022higher}. 

An issue related to these learning algorithms is the large amount of data necessary to train them properly. Regarding speed prediction, many of the works that focus on long and short term prediction use traffic sensor acquired data, thus using past speed values registered in a road to predict future ones. However, most of the roads do not have this historical speed data that can be used for time-series speed prediction. In some cases, there is no traffic data at all. The lack of data does not mean that the regions have no traffic, it just implies that we do not have a database that stores information from it. Another associated consequence is that when using these specific region data, even though there are works able to yield good predictions, in some cases we cannot transfer this same model to other regions. Therefore, to address the issue of non-existing speed historical data and the difficulty of applying the models to a different region, our work uses topographical road features to predict a vehicle's speed. We believe that the combination of common safety, law enforced driving behaviors (stopping at a traffic light), and road engineering (affecting a road's vehicle flow) have, in average, a wage on vehicle speed registrations. 

Therefore, by using a road's topographical features associated to GPS car trip information, we were able to train a model to predict a vehicle's speed given its trajectory. For this prediction, we based ourselves on a data point association scheme and a shared weights multilayer perceptron model (SWMLP). Three data association schemes are proposed, in which we create inputs for our model from a combination of past and future trajectory points. Based on our experiments, these inputs are capable of introducing to the model a sense of locality and the influence that one spatial point has on other's speed. The advantages of our model is the fact that when inferring predictions, we only need as inputs easily acquired topographical road features instead of historical speed data. We demonstrate in our results that it is possible after training to use the model in other geographical regions that might not have historical speed registrations. 

This paper is organized as follows. In Section~\ref{sec:method}, our proposed method is explained, and important concepts used in this work are clarified. In Section~\ref{sec:exp}, we describe the experiments performed and compare the achieved results to other standard methods. Finally, some concluding notes and suggestions for future work are presented in Section~\ref{sec:conc}.

\section{Proposed Method}
\label{sec:method}

\begin{figure}
    \centering
    \includegraphics[height=2.5 cm]{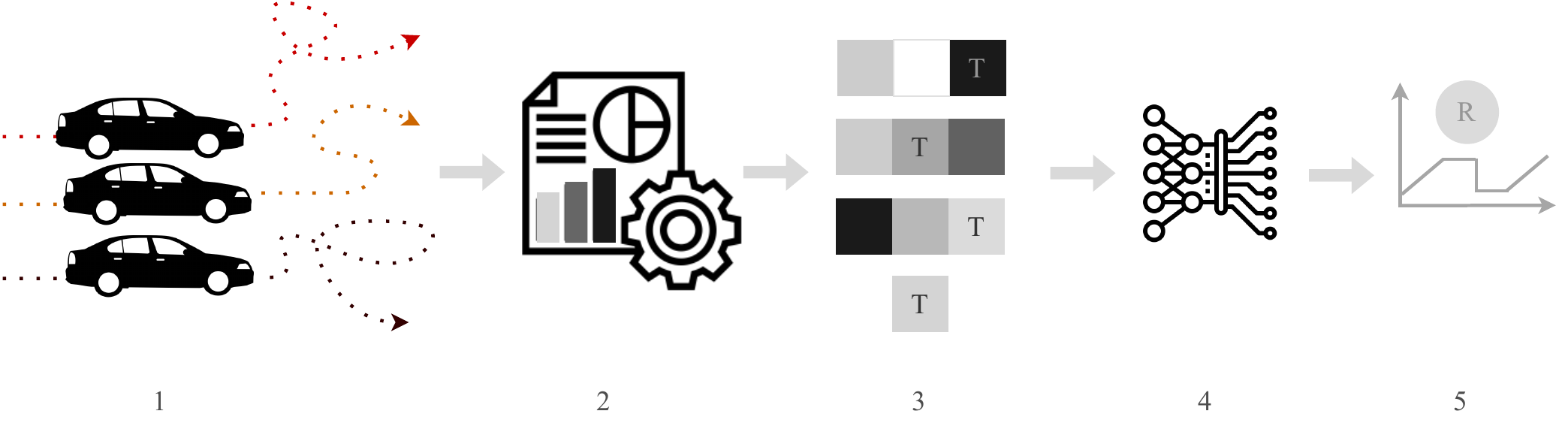}
    \caption{Methodology pipeline. (1) Use Geco air mobile application to collect car trips in a region, and HERE database to acquire topographical data from the region; (2) Process the data to extract viable features for speed prediction; (3) Associate the data points and generate inputs used for training, validation, and testing sets; (4) Train the network models; and (5) analyse predictions.}
    \label{fig:pipeline}
\end{figure}

The objective behind this work is to predict speed all along a given vehicle trajectory, prediction is based on the road topographical aspects. A local speed prediction differs from a short or long-term traffic prediction in a sense that it will be the speed prediction for a given data point and not a set of future points. To better understand this work some concepts need to be clarified: a vehicle's trip; the road topography; and a data point. In this work, a vehicle's trip is a GPS path registered in a database in which a car has a starting location and journeys, during a period of time, to a stopping location. We consider the topography, road features related to the architectural structure of the street such as traffic light, number of lanes, and the allowed speed limit, to name a few. Finally, a data point is a sampled GPS registration of a trip with the exact position of where the vehicle is and what road is it on. We opted to attend to this problem with no explicit time variables in the feature array. In this manner, it was possible to do an analysis of how predictions behaved without direct temporal elements. An overview of our work's pipeline can be observed in Figure~\ref{fig:pipeline} and will be discussed in the following subsections.

\subsection{Preprocessing}

\begin{figure}
    \centering
    \includegraphics[height=5cm]{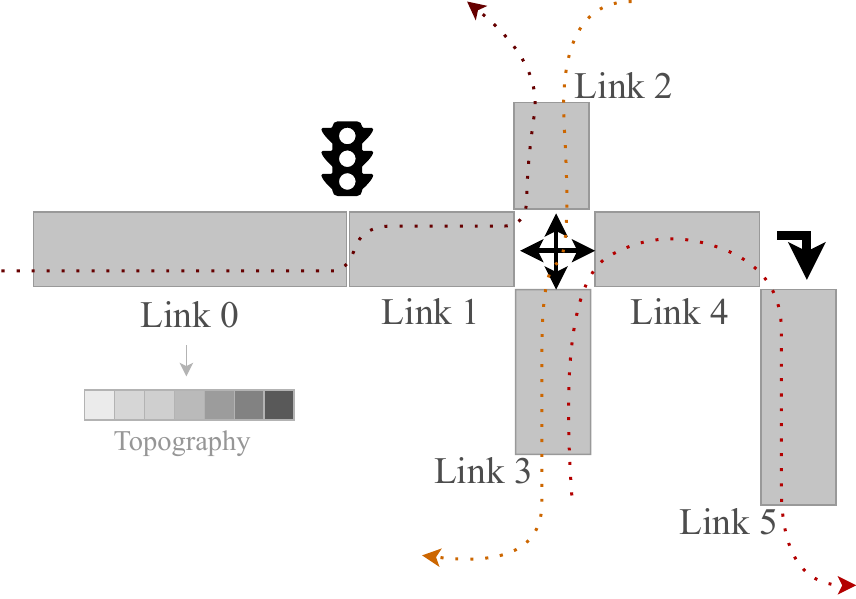}
    \caption{Road sections as links: Along the roads (represented by the rectangles), traffic lights, cross road, and other sort of cuts to the continuity are present. To have cleaner stored data with easier access to a specific road segment, a division is made (for the horizontal axis street of this image) in links 0, 1, and 4. Travel information: vehicles regularly drive on these maps utilizing multiple links in a single trip (dotted lines).}
    \label{fig:links}
\end{figure}

In this work we consider a road section as a link. This is done since an entire road might be comprised of multiple breakouts throughout its length such as intersections, and signalization (Figure~\ref{fig:links}). Thus, it is common to break down these roads in smaller more continuous sections transforming a road map into a link map. With these link maps, similarly to the roads, we are able to get vehicle travel information. 
\subsection{Features}

\begin{figure}
    \centering
    \includegraphics[width=\linewidth]{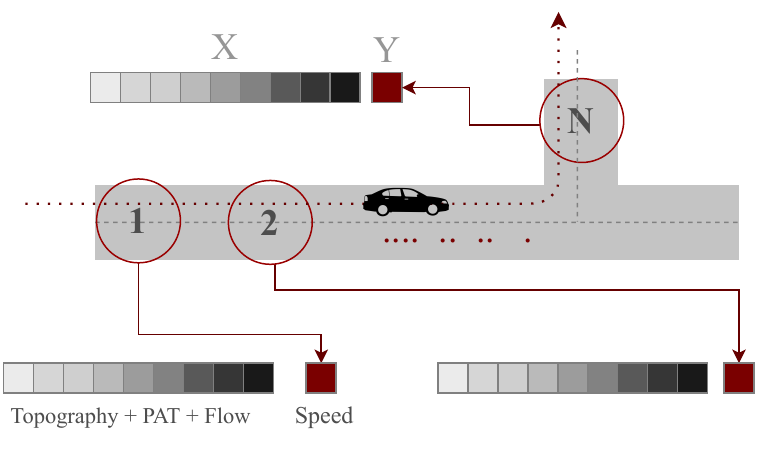}
    \caption{Data points we acquire for this work: A car trip has a set of multiple links associated to it, and at each GPS timestamp of the trip the car is at a different location (in the same link or not). For each trip we have $1$ to $N$ registered points where each of them have a set of topographical features (based on the link), and the speed registration.}
    \label{fig:dataPoints}
\end{figure}

As data points, we get the topographical features and their catalogued speed of every GPS registration of a car trip. This is illustrated in Figure~\ref{fig:dataPoints}. Although we get features based on GPS registration timesteps, we convert time into distance, thus time is not used as a feature.

We compiled eight features deemed to represent the basic road topography for each of the mentioned data points:

\begin{enumerate}
    \item link priority; 
    \item existence of traffic light at the beginning of the link; 
    \item existence of traffic light at the end of the link; 
    \item number of connected neighbor links; 
    \item number of lanes in a link; 
    \item speed limit; 
    \item link length; 
    \item percentage of already traveled link. 
\end{enumerate}

Link priority is a type of variable able to describe a level of importance to a street section, meaning that links with a higher demand (e.g. highway sections) have also higher values of priority. The existence of traffic light confirms or not if a link contains in one or both extremities a stop light. The connected neighbors are the number of possible different other links a car can access before or after entering the current section. Finally, to introduce variability to the feature vector creation, we calculate the percentage of already traveled (PAT) link. This feature represent the point in space a car is on a link. To better explain this we can consider a car driving through a link of length 50m, at timestep ($ts$) 2 the simulation placed the car at 7m, thus the PAT value would be \(PAT_{ts_2} = (7m\times100\%)/50m = 14\%\).

\subsection{Prediction schemes}\label{sec:schemes}
\begin{figure}
    \centering
    \includegraphics[height=3.5cm]{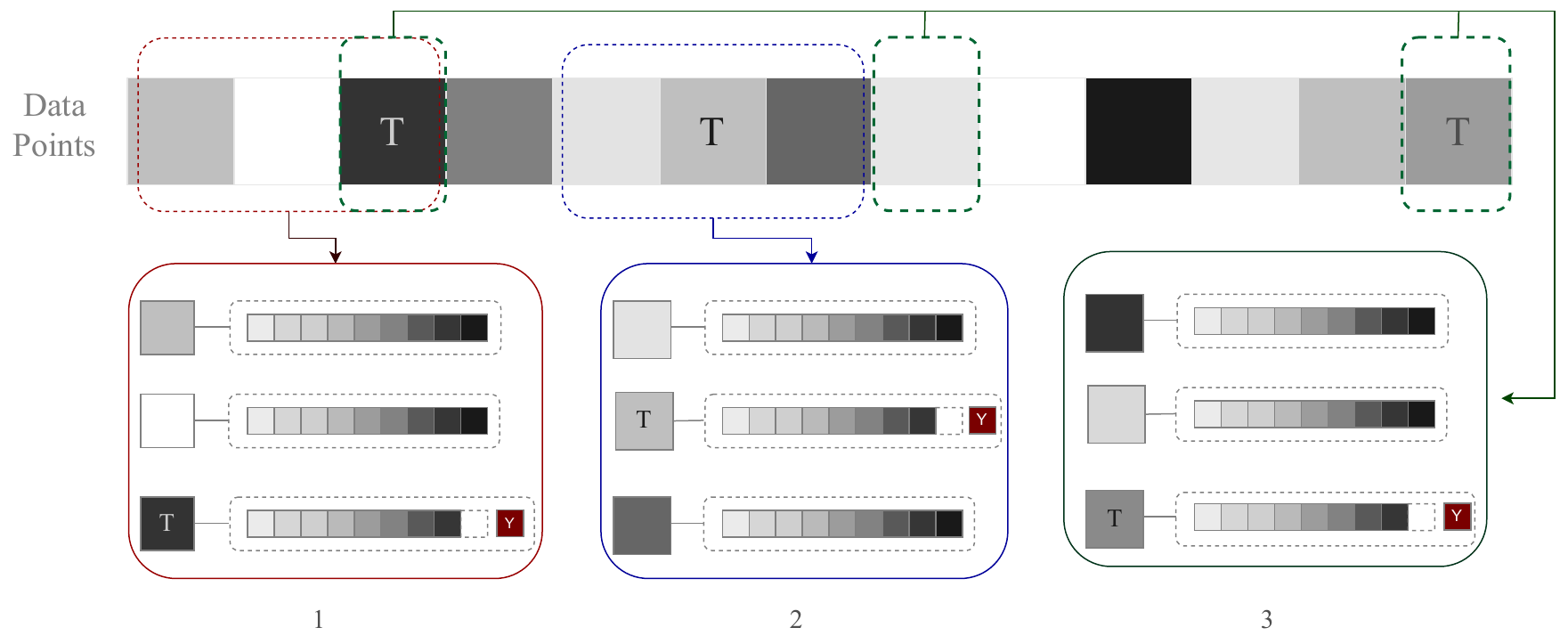}
    \caption{Three data point generation schemes for the SWMLP. After selecting the target data point, $T$: (1) \emph{Past} correlates the target to the features of the 2 most recent past data points; (2) \emph{Past and Future} joins the immediate past and future data point to the target; and (3) the \emph{Punctual Past} similar to 1 but a distance is allowed for the connected past points.}
    \label{fig:target}
\end{figure}

In this work, before prediction, we already have predefined trips with starting and ending location of the vehicle, as well as the path the vehicle took to arrive from start to end. Thus, to generate the inputs for our prediction model, we decided on the three association schemes, illustrated in Figure~\ref{fig:target}, as follows: 
\begin{enumerate}
\item \emph{Past} ($Pa$). We join in the same input three consecutive data points: a target and the two immediately previous registered data points. In other words, the target value will have its speed predicted based on its association with its two nearest consecutive topographical past points. 
\item \emph{Past and Future} ($P\&F$). We generate examples by associating the nearest past and future point topographical values to the target's, meaning that the prediction point is the middle data point.
\item \emph{Punctual Past} ($PuP$). Data points are collected in the past at a distance $d$ from the target. We used 5 and 10 GPS timesteps away.
\end{enumerate}

It is important to notice that for the targeted prediction points we do not use the PAT feature mentioned in the previous subsection. We wanted to verify, without the use of an explicit historical speed and no indication of where the vehicle will be, if we would be able to predict the target data point's speed.

\begin{figure}
    \centering
    \includegraphics[width=\linewidth]{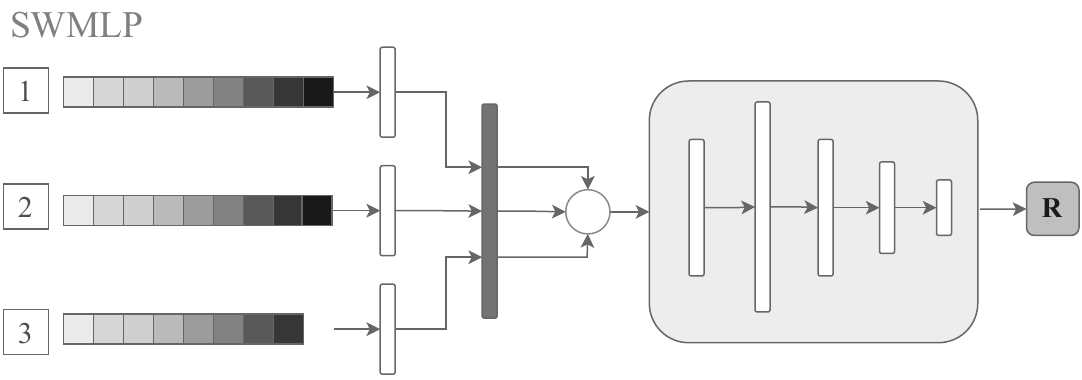}
    \caption{Representation of the SWMLP structure: we join three related data points, one target (3) and two neighbor acquired data point registration (1 and 2). In the targeted point features we do not consider the PAT feature so that it does not influence in prediction. For prediction, these points are embedded, passed through a shared weight layer, concatenated, and finally passed through an MLP structure.}
    \label{fig:method}
\end{figure}

 The main idea behind the SWMLP was to combine neighboring data points. By using $P\&F$, $Pa$, or $PuP$ data point associations, we are able to introduce to the network a sense of why and how a vehicle could change behavior in average. Considering situations where paths are defined, to validate a $P\&F$ data point association we contemplated the fact that the network might be able to understand that the vehicle will need to changed significantly the previous driving speed conditions to comply with the future point topographical features. For $Pa$, inspired by the concept of time-series, the network can implicitly learn, based on previous data points, an average local speed. Similar to $Pa$, we consider $PuP$ to verify if the distant past data point association can be better than an immediately previous one.  We opted to use a neural network as the prediction model. As our main contribution, we devised a Shared-Weight Multilayer Perceptron (SWMLP)(Figure~\ref{fig:method}) composed of: a layer of three individual linear layers (32 neurons each); a shared weight layer (16 neurons); and five layers (64, 128, 64, 32, 16 neurons).

In Equations~\ref{eq:1} to \ref{eq:4} we show how our model regresses. After assembling the inputs as sets of three points (either $Pa$, or $P\&F$, or $PuP$), the SWMLP divides each input in three streams. Each stream is fed to an embedding linear layer that will encode the individual features in a vector (Equation~\ref{eq:1}). This encoding is previously done so patterns between features in the same data point can be calculated before joining the streams.

\begin{equation}
\begin{aligned}
& f_{1} : \mathbf{x_{n}} \in \mathbb{R}^{d} \mapsto \mathbf{x'_{n}} \in \mathbb{R}^{s} \qquad f_{1}(\mathbf{x_{1}}) = \mathbf{x'_{1}},  \\
& f_{2} : \mathbf{x_{n}} \in \mathbb{R}^{d} \mapsto \mathbf{x'_{n}} \in \mathbb{R}^{s} \qquad f_{2}(\mathbf{x_{2}}) = \mathbf{x'_{2}}, \\
& f_{3} : \mathbf{x_{n}} \in \mathbb{R}^{d} \mapsto \mathbf{x'_{n}} \in \mathbb{R}^{s} \qquad f_{3}(\mathbf{x_{3}}) = \mathbf{x'_{3}}.  \\
\end{aligned}
\label{eq:1} 
\end{equation}

We input these embeddings to same linear layer that will share the weights learned between these inputs (Equation~\ref{eq:2}). Weights of these three streams are thus shared and jointly updated. 
\begin{equation}
\begin{aligned}
& g : x'_{n} \in \mathbb{R}^{s} \mapsto x''_{n} \in \mathbb{R}^{s'}, \\
& g(x'_{1}) = x''_{1}, \qquad  g(x'_{2}) = x''_{2}, \qquad  g(x'_{3}) = x''_{3}.
\end{aligned}
\label{eq:2}
\end{equation}

Following the weight sharing, we concatenate the layer outputs (Equation~\ref{eq:3}) and use them as inputs for a five layered MLP (Equation~\ref{eq:4}). 

\begin{equation}
X = \big[{x''_{1}}^\top \, {x''_{2}}^\top \, {x''_{3}}^\top\big]^\top,
\label{eq:3}   
\end{equation}

\begin{equation}
\begin{aligned}
& \textrm{Reg} : X_{n} \in \mathbb{R}^{s''} \mapsto X'_{n} \in \mathbb{R}^{1}, \\
& \textrm{Reg}(X) = p.
\end{aligned}
\label{eq:4}   
\end{equation}

To further assure that time does not explicitly influence the regression results, we shuffle the inputs to remove an explicit trajectory order inside the training and validation set.

\section{Experiments}
\label{sec:exp}

Experiments were conducted to verify that our SWMLP model (presented in Section~\ref{sec:method}) was able to reconstruct and thus predict the speed of a predetermined car trajectory based on the topographical information of a link. We used real-world data collected from IFPEN's mobile application Geco air\footnote{https://www.gecoair.fr/en/}, and HERE\footnote{https://www.here.com} databases. From these databases, we gather car trip data as well as obtain the topographical characteristics of the driven links. 

Our model was compared to three standard regression algorithms: Multilayer Perceptron (MLP), Support Vector Regression (SVR), and Random Forest (RF). These models work with each data point as an individual input; each feature vector described in Figure~\ref{fig:dataPoints} is associated to one speed value, without combining the neighboring feature vectors. The MLP is composed of the same final five layers of the SWMLP, making it the most directly comparable to our approach. All the other hyper-parameters in MLP, SVR, and RF were determined by cross-validation. We conducted experiments with these standart algorithms because we wanted to verify if the individual data point, without association was able to be used by itself in prediction. Since we do not have data point association in these standard experiments we allow the use of the PAT feature for the single target.

Three link map scenarios were selected: (i) Nanterre; (ii) Paris; (iii) Lyon. The three scenarios were used individually to generate data for a training, validation, and testing sets, respectively. The training set (Nanterre) contained 440'550 data points collected from multiple car trips from January to May of 2018. The validation set (Paris) was devised so that the number of final data points was 30\% of the size of the training set, hence, it had 132'000 points. And finally, the test set (Lyon) was generated from collecting trip data from 1 month during January of 2018, leading to 630 complete trips. Although being collected at the same time as training and validation, since we select a different region to create the test there is no overlapping data between them. We chose to build our test set from a further region compared to training and validation to analyse the robustness of the predictions when dealing with different cities, thus, have a divergence of topographical feature combinations. In addition, differently from training and validation sets, we do not shuffle the test set since we need the correct sequence of data points to perform the prediction inference.

\begin{figure}
    \centering

    \includegraphics[height=9.7cm]{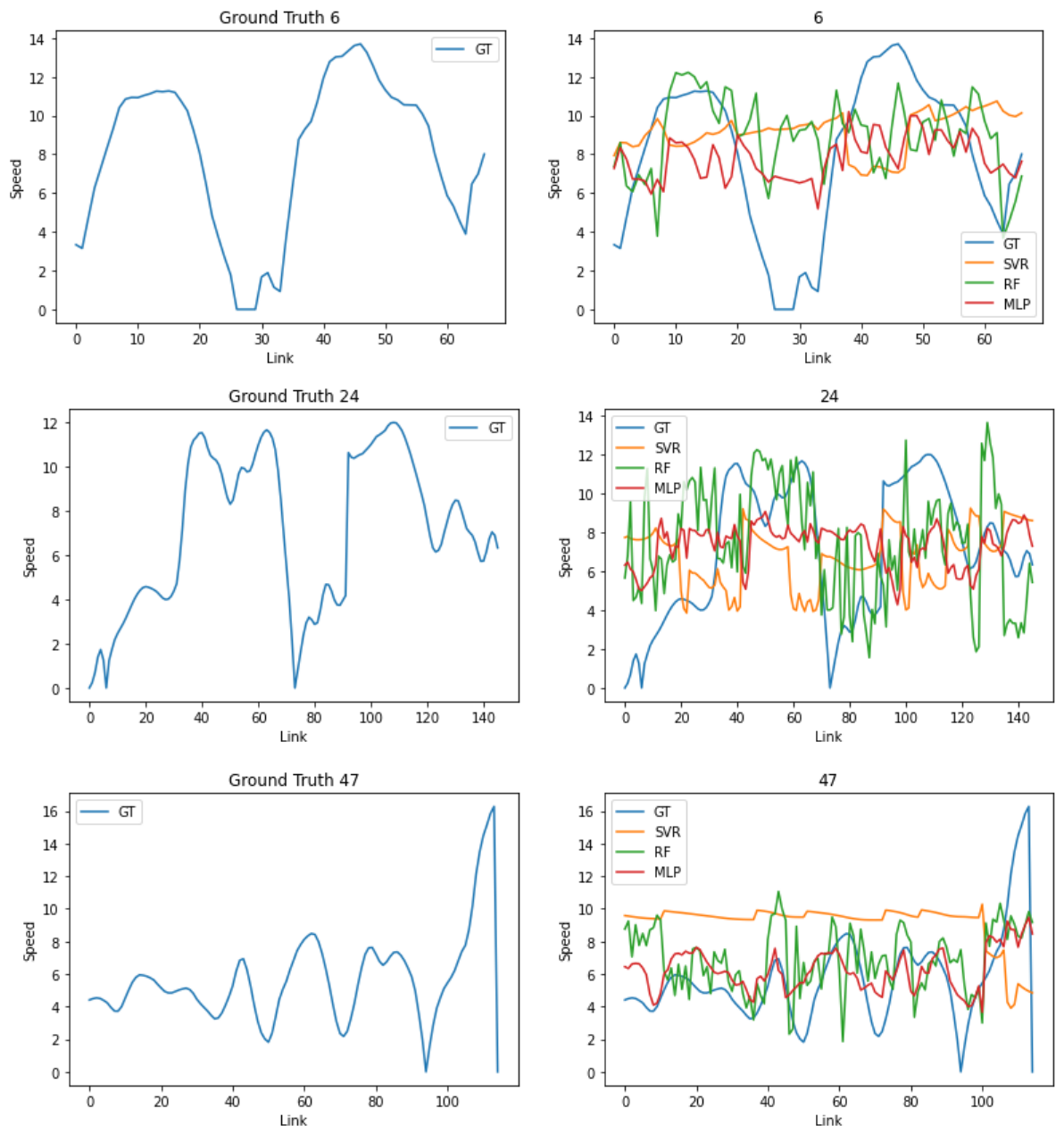}

    \caption{Three randomly chosen trips (labeled 6, 24, and 47). First column: True speed values for individual trips. Second column: Speed predictions of the $MLP$ baseline, $SVR$, and Random Forest $RF$ regression algorithms.}
    \label{res:base}
\end{figure}

Concerning a qualitative analysis, in Figure~\ref{res:base} we can see prediction results of the SVR, Random Forest, and MLP baselines. For this training we conducted a grid search over the parameters to best fit each algorithm to the train data. It is possible to observe from the graphs that results from all three algorithms were noisy. However, with only the topographical features from the single GPS timestep instance data points they were able to learn and follow a coherent speed prediction mean. It is possible to consider the fact that individual data point regression, in this configuration, poorly regresses speed. Even when the exact location of where the vehicle is in the link is disclosed, the baselines still appears to struggles to yield consistent results compared to the ground truth.

\begin{figure}
    \centering
    \includegraphics[height=9.7cm]{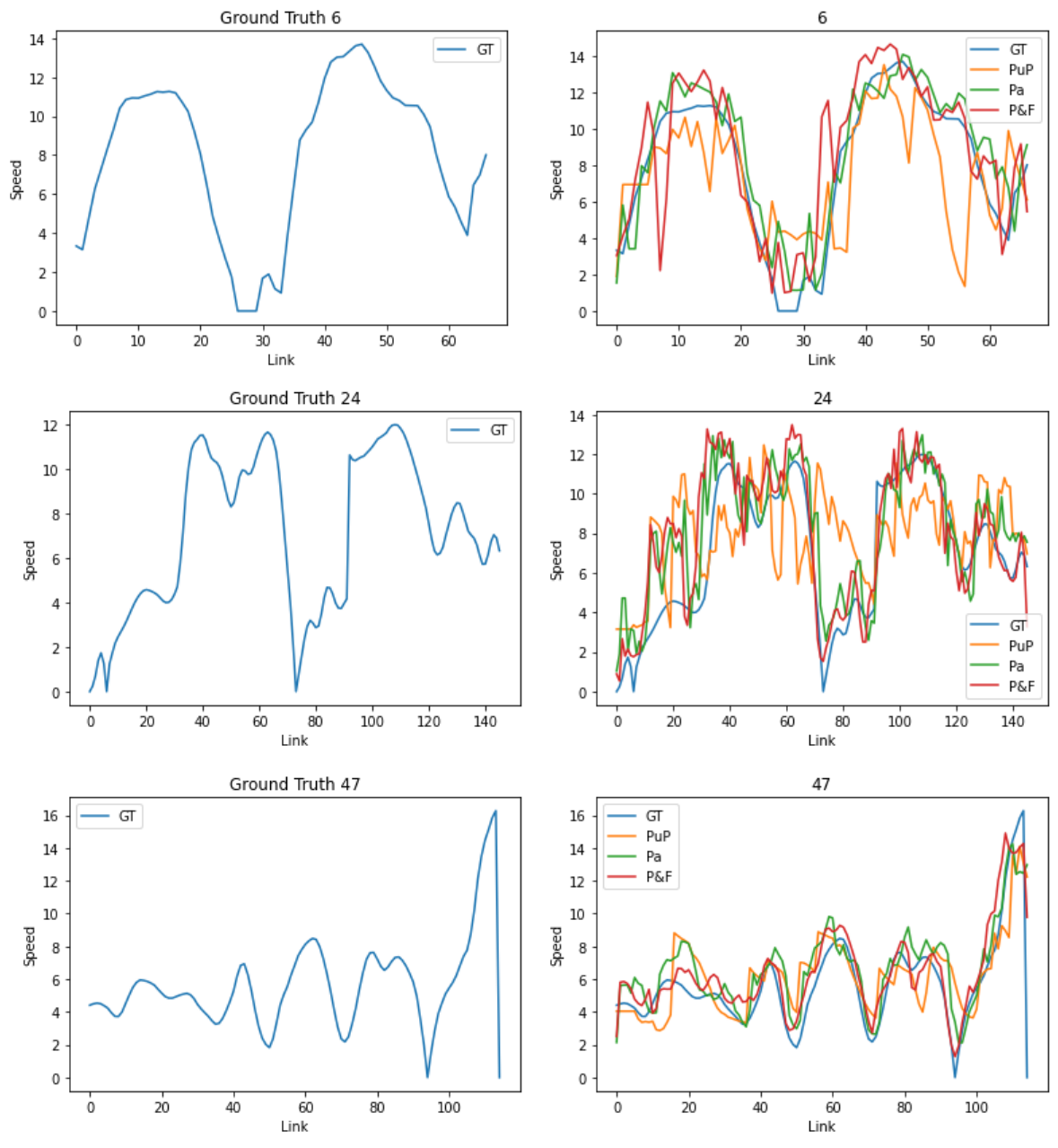}
    \caption{Different trip (labeled 6, 24, and 47) prediction graphs: First column: True speed values for individual trips. Second column: Speed prediction of the proposed SWMLP with $Pa$, $P\&F$, or $PuP$ scheme. The Y coordinate represent speed values and the X represent the ordered data point sequence that compose a full trip.}
    \label{res:sw}
\end{figure}

Figure~\ref{res:sw} shows a comparison of our three prediction schemes presented in Section \ref{sec:schemes} ($Pa$, $P\&F$, and $PuP$). We compare the three predictions for the same ground truths presented in Figure~\ref{res:base}. Even if qualitatively results are similar to the ground truth, the best predictions are yielded when using $P\&F$ and $Pa$. For the $P\&F$, in addition to knowing where the car was, we assume that the knowledge of its future spatial configuration brings further information to the regression model. This means that with the information of a sudden change in the car's trajectory, the network can better preview what would be the target's speed value. It is also deducible that the reason prediction is more unstable by using the $PuP$ association is because with larger spatial gaps between the target and associated points (5 and 10 points further), we lose some information for immediate changes in speed. However, $PuP$ can still yield better predictions compared to the isolated data point predictions of the standard regressors. The balance between these, the $Pa$ data point association, also generated compatible results to the ground truth. 

\begin{figure}[!t]
    \centering
    \includegraphics[height=15cm]{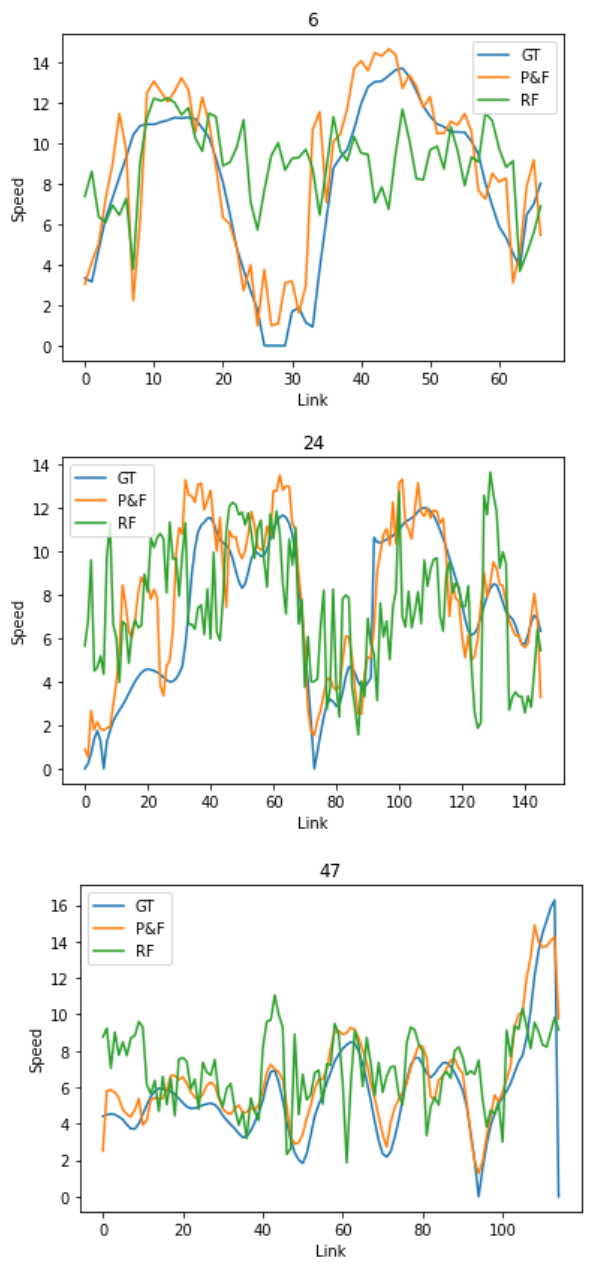}
    \caption{Three different trip (labeled 6, 24, and 47) prediction graphs: Qualitative comparison between Random Forest ($RF$) regression algorithm and our $P\&F$ approach. The Y coordinate represent speed values and the X represent the ordered data point sequence that compose a full trip.}
    \label{res:comp}
\end{figure}

To avoid a noisy an unreadable graph comparison and since the baselines did not show a significant qualitative difference between them, we chose only the Random Forest results to compare it to our $P\&F$ approach in Figure~\ref{res:comp}. We can see that even though $P\&F$ also contains noise, it is better fitted to the ground truth compared to the Random Forest. Another remark is the ability of $P\&F$ to match the ground truth valleys compared to the other regressors.

Regarding a quantitative analysis, our speed was calculated in meters per seconds and we can compare the metrics for the regression algorithms in Table~\ref{tab:comp}. Between the baseline algorithms, MLP had the best values in average for the predictions while testing; however, Random Forest and SVR did not differ much from MLP. For our proposed approach, with all of the different data point associations, we can see that we were able to significantly reduce the values of error compared to MLP, Random Forest, and SVR. As regards to the quantitative comparison between our data point association we can see that even though qualitatively $P\&F$ and $Pa$ had similar predictions, when dealing with metric values $P\&F$ had a superior performance.

\begin{table}
\caption{Quantitative comparison between regression algorithms. The presented metrics were: Rooted Mean Squared Error, Mean Squared Error, and Mean Absolute Error. These metrics were averaged over the individual metrics calculated over all test set and not only the presented qualitative ones.}
\centering
\begin{tabular}{lccc}
\multicolumn{1}{c}{} & RMSE & MSE    & MAE \\ \hline
SVR                  & 5.25 & 32.25 &  4.52    \\ \hline
RF                   & 4.61 & 23.93  & 3.86    \\ \hline
MLP                  & 4.29 & 23.21  & 3.69   \\ \hline

SWMLP ($PuP$)        &  3.75         &     16.93       & 3.13 \\ \hline
SWMLP ($Pa$)         &  2.77         &     9.81        & 2.15   \\ \hline
SWMLP ($P\&F$)       & \textbf{2.57} & \textbf{8.67}   & \textbf{1.94}
\end{tabular}

\label{tab:comp}
\end{table}

\section{Conclusion}
\label{sec:conc}

In this work we devised a method for predicting a car's trajectory speed based on topographical features and data point association. Here, we prove that we are able to use the road's topographical features to generate the speed prediction on a trajectory. The advantages of this approach is that by acquiring the trajectory and having the topographical map features of its component links, we do not need time series data to generate an average behavior. Given that this work is able to predict a corresponding speed dependant on predetermined paths, it is also our intention design a solution so that this prediction becomes more independent. Path dependency signifies that a link's speed can vary according to the modification in a car's trajectory, thus we will work on the prediction of a range of possible speed values for specific a data point as well as create different data point associations schemes able to convey the idea of path generalization. Finally, it is also our intention to continue working on trajectory speed prediction and smooth out the noise of our current prediction, and include in our features the use of previous speed predicted values to verify if changes are significant in the results.

\bibliographystyle{IEEEbib}
\bibliography{refs}

\end{document}